\documentclass[10pt,journal,compsoc]{IEEEtran}

\ifCLASSOPTIONcompsoc
  \usepackage[nocompress]{cite}
\else
  \usepackage{cite}
\fi

\ifCLASSINFOpdf
\else
\fi

\usepackage{graphicx}
\usepackage{amsfonts}
\usepackage{amsmath}
\usepackage{color,soul}
\usepackage[hidelinks]{hyperref}

\usepackage{enumitem, color, amssymb}
\newlist{inlinelist}{enumerate*}{1}
\setlist*[inlinelist,1]{%
	label=(\roman*),
	before=\unskip{: }, itemjoin={{, }}, itemjoin*={{, and }}
}

\usepackage{algorithm}
\usepackage[noend]{algpseudocode}
\begin{document}
	
\title{Reasoning with shapes: profiting cognitive susceptibilities to infer linear mapping transformations between shapes}

\author{Vahid~Jalili
\IEEEcompsocitemizethanks{\IEEEcompsocthanksitem Oregon Health \& Science University, Portland, OR, USA}}

\markboth{Reasoning with Shapes}%
{Reasoning with Shapes}

\IEEEtitleabstractindextext{%
\begin{abstract}
Visual information plays an indispensable role in our daily interactions with environment. Such information is manipulated for a wide range of purposes spanning from basic object and material perception to complex gesture interpretations. There have been novel studies in cognitive science for in-depth understanding of visual information manipulation, which lead to answer questions such as: how we infer 2D/3D motion from a sequence of 2D images? how we understand a motion from a single image frame? how we see forest avoiding trees?

Leveraging on congruence, linear mapping transformation determination between a set of shapes facilitate motion perception. Present study methodizes recent discoveries of human cognitive ability for scene understanding. The proposed method processes images hierarchically, that is an iterative analysis of scene abstractions using a rapidly converging heuristic iterative method. The method hierarchically abstracts images; the abstractions are represented in polar coordinate system, and any two consecutive abstractions have incremental level of details. The method then creates a graph of approximated linear mapping transformations based on circular shift permutations of hierarchical abstractions. The graph is then traversed in best-first fashion to find best linear mapping transformation. The accuracy of the proposed method is assessed using normal, noisy, and deformed images. Additionally, the present study deduces (i) the possibility of determining optimal mapping linear transformations in logarithmic iterations with respect to the precision of results, and (ii) computational cost is independent from the resolution of input shapes. 

\end{abstract}

}

\maketitle
\IEEEdisplaynontitleabstractindextext
\IEEEpeerreviewmaketitle

\IEEEraisesectionheading{\section{Introduction}}

\IEEEPARstart{V}{ision} system is studied in orthogonal disciplines spanning from neurophysiology and psychophysics to computer science all with uniform objective: understand the vision system and develop it into an integrated theory of vision. In general, vision or visual perception is the ability of information acquisition from environment, and it's interpretation. According to Gestalt theory, visual elements are perceived as patterns of wholes rather than the sum of constituent parts~\cite{koffka2013principles}. The Gestalt theory through \textit{emergence}, \textit{invariance}, \textit{multistability}, and \textit{reification} properties (aka Gestalt principles), describes how vision recognizes an object as a \textit{whole} from constituent parts. There is an increasing interested to model the cognitive aptitude of visual perception; however, the process is challenging. In the following, a challenge (as an example) per object and motion perception is discussed.

\subsection{Why do things look as they do?}
In addition to Gestalt principles, an object is characterized with its spatial parameters and material properties. Despite of the novel approaches proposed for material recognition (e.g.,~\cite{sharan2013recognizing}), objects tend to get the attention. Leveraging on an object's spatial properties, material, illumination, and background; the mapping from real world 3D patterns (distal stimulus) to 2D patterns onto retina (proximal stimulus) is many-to-one non-uniquely-invertible mapping~\cite{dicarlo2007untangling,horn1986robot}. There have been novel biology-driven studies for constructing computational models to emulate anatomy and physiology of the brain for real world object recognition (e.g.,~\cite{lowe2004distinctive,serre2007robust,zhang2006svm}), and some studies lead to impressive accuracy. For instance, testing such computational models on gold standard controlled shape sets such as Caltech101 and Caltech256, some methods resulted $<$60\% true-positives~\cite{zhang2006svm,lazebnik2006beyond,mutch2006multiclass,wang2006using}. However, Pinto et al.~\cite{pinto2008real} raised a caution against the pervasiveness of such shape sets by highlighting the unsystematic variations in objects features such as spatial aspects, both between and within object categories. For instance, using a V1-like model (a neuroscientist's null model) with two categories of systematically variant objects, a rapid derogate of performance to 50\% (chance level) is observed~\cite{zhang2006svm}. This observation accentuates the challenges that the infinite number of 2D shapes casted on retina from 3D objects introduces to object recognition. 

Material recognition of an object requires in-depth features to be determined. A mineralogist may describe the luster (i.e., optical quality of the surface) with a vocabulary like greasy, pearly, vitreous, resinous or submetallic; he may describe rocks and minerals with their typical forms such as acicular, dendritic, porous, nodular, or oolitic. We perceive materials from early age even though many of us lack such a rich visual vocabulary as formalized as the mineralogists~\cite{adelson2001seeing}. However, methodizing material perception can be far from trivial. For instance, consider a chrome sphere with every pixel having a correspondence in the environment; hence, the material of the sphere is hidden and shall be inferred implicitly~\cite{shafer2000color,adelson2001seeing}. Therefore, considering object material, object recognition requires surface reflectance, various light sources, and observer's point-of-view to be taken into consideration.

\subsection{What went where?}
Motion is an important aspect in interpreting the interaction with subjects, making the visual perception of movement a critical cognitive ability that helps us with complex tasks such as discriminating moving objects from background, or depth perception by motion parallax. Cognitive susceptibility enables the inference of 2D/3D motion from a sequence of 2D shapes (e.g., movies~\cite{niyogi1994analyzing,little1998recognizing,hayfron2003automatic}), or from a single image frame (e.g., the pose of an athlete runner~\cite{wang2013learning,ramanan2006learning}). However, its challenging to model the susceptibility because of many-to-one relation between distal and proximal stimulus, which makes the local measurements of proximal stimulus inadequate to reason the proper global interpretation. One of the various challenges is called \textit{motion correspondence problem}~\cite{attneave1974apparent,ullman1979interpretation,ramachandran1986perception,dawson1991and}, which refers to recognition of any individual component of proximal stimulus in frame-1 and another component in frame-2 as constituting different glimpses of the same moving component. If one-to-one mapping is intended, $n!$ correspondence matches between $n$ components of two frames exist, which is increased to $2^n$  for one-to-any mappings. To address the challenge, Ullman~\cite{ullman1979interpretation} proposed a method based on nearest neighbor principle, and Dawson~\cite{dawson1991and} introduced an auto associative network model. Dawson's network model~\cite{dawson1991and} iteratively modifies the activation pattern of local measurements to achieve a stable global interpretation. In general, his model applies three constraints as it follows
\begin{inlinelist}
	\item \textit{nearest neighbor principle} (shorter motion correspondence matches are assigned lower costs)
	\item \textit{relative velocity principle} (differences between two motion correspondence matches)
	\item \textit{element integrity principle} (physical coherence of surfaces)
\end{inlinelist}.
According to experimental evaluations (e.g.,~\cite{ullman1979interpretation,ramachandran1986perception,cutting1982minimum}), these three constraints are the aspects of how human visual system solves the motion correspondence problem. Eom et al.~\cite{eom2012heuristic} tackled the motion correspondence problem by considering the relative velocity and the element integrity principles. They studied one-to-any mapping between elements of corresponding fuzzy clusters of two consecutive frames. They have obtained a ranked list of all possible mappings by performing a state-space search.

\subsection{How a stimuli is recognized in the environment?}

Human subjects are often able to recognize a 3D object from its 2D projections in different orientations~\cite{bartoshuk1960mental}. A common hypothesis for this \textit{spatial ability} is that, an object is represented in memory in its canonical orientation, and a \textit{mental rotation} transformation is applied on the input image, and the transformed image is compared with the object in its canonical orientation~\cite{bartoshuk1960mental}. The time to determine whether two projections portray the same 3D object
\begin{inlinelist}
	\item increase linearly with respect to the angular disparity~\cite{bartoshuk1960mental,cooperau1973time,cooper1976demonstration}
	\item is independent from the complexity of the 3D object~\cite{cooper1973chronometric}
\end{inlinelist}.
Shepard and Metzler~\cite{shepard1971mental} interpreted this finding as it follows: \textit{human subjects mentally rotate one portray at a constant speed until it is aligned with the other portray.}

\subsection{State of the Art}

The linear mapping transformation determination between two objects is generalized as determining optimal linear transformation matrix for a set of observed vectors, which is first proposed by Grace Wahba in 1965~\cite{wahba1965least} as it follows. 
\textit{Given two sets of $n$ points $\{v_1, v_2, \dots v_n\}$, and $\{v_1^*, v_2^* \dots v_n^*\}$, where $n \geq 2$, find the rotation matrix $M$ (i.e., the orthogonal matrix with determinant +1) which brings the first set into the best least squares coincidence with the second. That is, find $M$ matrix which minimizes}
\begin{equation}
	\sum_{j=1}^{n} \vert v_j^* - Mv_j \vert^2
\end{equation}

Multiple solutions for the \textit{Wahba's problem} have been published, such as Paul Davenport's q-method. Some notable algorithms after Davenport's q-method were published; of that QUaternion ESTimator (QU\-EST)~\cite{shuster2012three}, Fast Optimal Attitude Matrix \-(FOAM)~\cite{markley1993attitude} and Slower Optimal Matrix Algorithm (SOMA)~\cite{markley1993attitude}, and singular value decomposition (SVD) based algorithms, such as Markley’s SVD-based method~\cite{markley1988attitude}. 

In statistical shape analysis, the linear mapping transformation determination challenge is studied as Procrustes problem. Procrustes analysis finds a transformation matrix that maps two input shapes closest possible on each other. Solutions for Procrustes problem are reviewed in~\cite{gower2004procrustes,viklands2006algorithms}. For orthogonal Procrustes problem, Wolfgang Kabsch proposed a SVD-based method~\cite{kabsch1976solution} by minimizing the root mean squared deviation of two input sets when the determinant of rotation matrix is $1$. In addition to Kabsch’s partial Procrustes superimposition (covers translation and rotation), other full Procrustes superimpositions (covers translation, uniform scaling, rotation/reflection) have been proposed~\cite{gower2004procrustes,viklands2006algorithms}. The determination of optimal linear mapping transformation matrix using different approaches of Procrustes analysis has wide range of applications, spanning from forging human hand mimics in anthropomorphic robotic hand~\cite{xu2012design}, to the assessment of two-dimensional perimeter spread models such as fire~\cite{duff2012procrustes}, and the analysis of MRI scans in brain morphology studies~\cite{martin2013correlation}.

\subsection{Our Contribution}

The present study methodizes the aforementioned mentioned cognitive susceptibilities into a cognitive-driven linear mapping transformation determination algorithm. The method leverages on mental rotation cognitive stages~\cite{johnson1990speed} which are defined as it follows
\begin{inlinelist}
	\item a mental image of the object is created
	\item object is mentally rotated until a comparison is made
	\item objects are assessed whether they are the same
	\item the decision is reported
\end{inlinelist}.
Accordingly, the proposed method creates hierarchical abstractions of shapes~\cite{greene2009briefest} with increasing level of details~\cite{konkle2010scene}. The abstractions are presented in a vector space. A graph of linear transformations is created by circular-shift permutations (i.e., rotation superimposition) of vectors. The graph is then hierarchically traversed for closest mapping linear transformation determination. 

Despite of numerous novel algorithms to calculate linear mapping transformation, such as those proposed for Procrustes analysis, the novelty of the presented method is being a cognitive-driven approach. This method augments promising discoveries on motion/object perception into a linear mapping transformation determination algorithm.

\section{Method}
\paragraph*{Basic manipulations vs. complex calculations}
An infant has intuitive understanding of numbers and shapes, and can distinguish numerical and identity invariance of objects~\cite{izard2008distinct} regardless of object domain~\cite{wynn2002enumeration}; an ability that surprisingly extends to non-object entities (e.g., action~\cite{sharon1998individuation}). It lets us argue that an infant has basic understanding of transformations by primary perception of numbers and shapes. This intuitive ability is based on an early development of \textit{approximate number system}. This ability encourages present study to concentrate on basic operations and visual properties of shapes for the linear mapping transformation determination task.

\paragraph*{Abstract vs. detailed representations}
Of the entire environment within our visual range, only the essential information for the action in progress is prominent and the rest of the details are ignored~\cite{intraub1997representation} (aka cognitive inhibition~\cite{macleod2007concept}).  For instance, while crossing a street only the information about the direction and speed of cars on the street are required; details such as plate number of the cars or clothes drivers wore, generally not consciously registered in the visual perception. This highlights the significant role that abstractions play in reducing the amount of information to be considered. Additionally, Ballard~\cite{ballard1997deictic} and Agre~\cite{agre1987pengi} further explained this ability as deictic strategies where eye fixation point is used to guide body movement (modeling the behavior at the embodiment level) while fixation point can rapidly change to different location~\cite{ballard1991animate}. 

In the following, we discuss how the proposed method abstracts images and determines linear mapping transformations between the images.

\subsection{Shape representation} \label{section: Shapre Representation}
The overall procedure of the presented method is independent from the color model of input shape (i.e., RGB, Cyan Magenta Yellow Key (CMYK), Hue Saturation Value (HSV), B\&W, binary, and etc.).
Present study manipulates binary representation of shapes; while extension to other color models is straightforward and requires the modification of segment aggregation function (discussed in Section~\ref{section: Shape Segmentation}). The motivations of binarizing shapes are threefold. First, simple aggregation functions such as count can be applied on binary shapes. This improves the readability of the presented method, and avoids various color-model-based aggregation functions, which are beyond the scope of this manuscript.

Second, real world objects incorporate spatial parameters and materials, introducing distal-to-proximal stimulus mapping challenges, and motion correspondence problem. The objectives of present study are to methodize principle cognitive susceptibilities for transformation determination, and the fact that binary shapes are not as sensible to aforementioned challenges as colorful shapes are, makes the binary model suitable for present study. The binary color model, is a common model among the motion correspondence problem studies (e.g.,~\cite{girod2013principles,hirschmuller2009evaluation}.

Third, a \textit{distance transform} and topological skeleton extraction from binary shapes is straightforward as opposed to colorful shapes. These transformations are applicable alternatives for the segment aggregation functions discussed in Section~\ref{section: Shape Segmentation}. Despite of the promising methods that exist in literature (e.g., topological volume skeletonization~\cite{takahashi2004topological}, or various methods on distance transform algorithms~\cite{fabbri20082d}), to best of our knowledge, none of the proposed methods are comprehensive in the consideration of full explanatory real world object characteristics such as material, illumination, and surface reflectance. For instance, the pattern in a chro\-mium plated sphere in an image frame is indeed reflecting the surrounding environment and the sphere itself is determined implicitly~\cite{adelson2001seeing}. However, binary shapes mask similar properties encouraging least ambiguity.

\begin{figure}[!t]
	\centering
	\includegraphics[width=0.95\columnwidth]{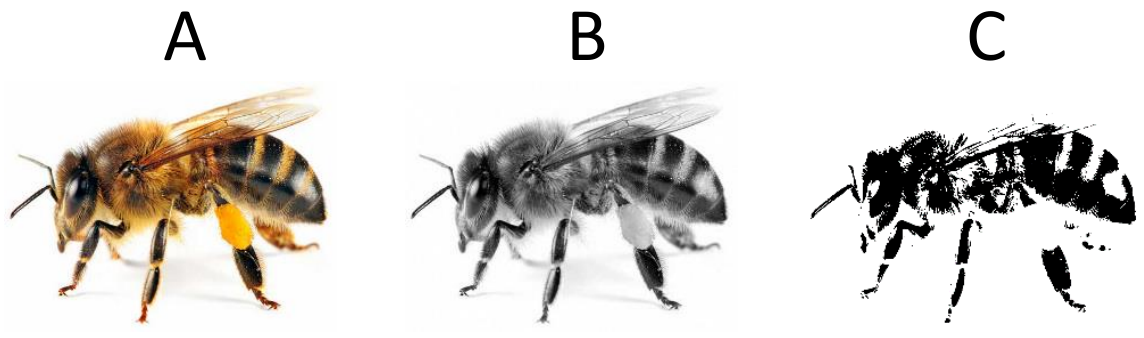}
	\caption
	{
		\textbf{A:} input shape in RGB color model; \textbf{B:} intermediate gray-scale representation of the input; \textbf{C:} binary representation of the input. The proposed method manipulates the binary representation.
	}
	\label{Figure: ShapeRepresentation}
\end{figure}

Present study manipulates binary shapes. In this regard, first a colored shape is converted to its corresponding gray-scale B\&W frame. The procedure is by estimating the luminance for every pixel $x$, $y$ of an image frame in RGB color model (e.g., panel A on Fig.~\ref{Figure: ShapeRepresentation}) as $L_{xy} = 0.2126R + 0.7152G + 0.0722B$ (see panel B on Fig.~\ref{Figure: ShapeRepresentation}). The resulted B\&W image frame is then binarized by normalizing $L_{xy}$ as $B_{xy} = \lfloor L_{xy} / 128 \rfloor$ for the binary pixel $B_{xy}$ (see panel C on Fig.~\ref{Figure: ShapeRepresentation}). Note that, $L_{xy} \in \{0, 1, \dots 255\}$, therefore, $B_{xy} \in \{0, 1\}$.

\subsection{Shape segmentation} \label{section: Shape Segmentation}
Shape segmentation is a well-studied subject in the field of image processing (e.g.,~\cite{pal1993review}). Segmentation commonly proceeds shape semantic analysis, thus highlighting the necessity of adapting segmentation method to the objectives of the study. Accordingly, the segmentation procedure is defined as it follows, which emphasizes the relative location of the pixels to facilitate liner mapping transformation.

\begin{figure}[!t]
	\centering
	\includegraphics[width=0.7\columnwidth]{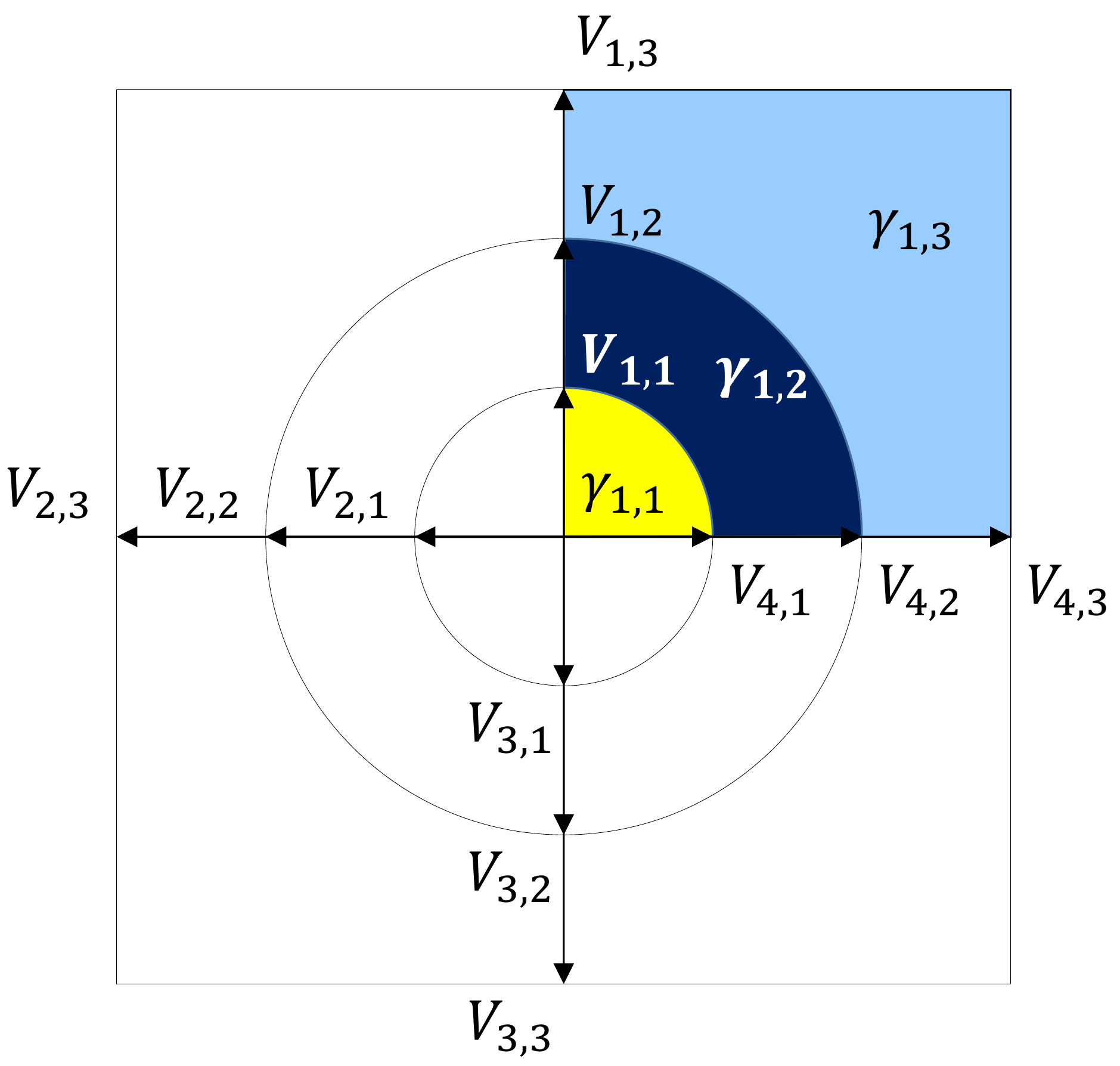}
	\caption
	{
		This shape illustrates segmentation with $N=4$ sectors and $M=3$ segments. A segment represents an area of the frame. For instance, $\vec{V}_{1,1}$, $\vec{V}_{1,2}$, and $\vec{V}_{1,3}$ respectively represent the areas shaded in yellow, blue, and cyan. Each area is composed of a set of pixels aggregated in $\gamma_{nm}$. For instance, $\gamma_{1,1}$, $\gamma_{1,2}$, and $\gamma_{1,3}$ denote the aggregation of pixels located at yellow, green, and blue shaded areas respectively.
	}
	\label{Figure: ShapeSegmentation}
\end{figure}

In present study, shapes are considered in two Dimensional (2D) Euclidean space. An image frame is segmented in $N$ \textit{sectors} and $M$ \textit{segments} (see Fig.~\ref{Figure: ShapeSegmentation}). Sectors are divisions of the image frame in \textit{angle} ($\varphi$) direction of polar coordinate system and are denoted by Euclidean unit vector $\vec{V}_n$ for $n\in\{1,2, \dots N\}$. Segments are isometric divisions of sectors in \textit{radius} ($r$) direction of polar coordinate system, denoted by the Euclidean unit vector $\vec{V}_{nm}$ for $m \in \{1,2, \dots M\}$. Accordingly, all sectors have equal number of segments, and segments are the smallest segmentation units. Let $I$ denote \textit{segmentation matrix} defined as it follows.
\[
\mathbf{I} =
\bordermatrix{ & \text{Segment} \; 1             & \dots & \text{Segment} \; M  \cr
	  \text{Sector} \; 1 & \vec{V}_{11} & \dots  & \vec{V}_{1M}       \cr
	     \dots & \vdots       & \ddots & \vdots             \cr
	  \text{Sector} \; N & \vec{V}_{N1} & \dots  & \vec{V}_{NM}}      \qquad
\]
\noindent
Each element $V_{nm}$ is a tuple of $\langle x, y, \gamma \rangle$, where $\gamma$ is an aggregated value of a portion of the image frame which is represented by $V_{nm}$. Note that, the dimension of segmentation matrix is independent from the resolution of input image frame. 

The area represented by a segment $\vec{V}_{nm}$ is characterized by two boundaries environing it, and it is defined in polar coordinate system as it follows.
\begin{align} 
r \in & \left] \frac{m-1}{M}       , \frac{m}{M}    \right] \\ 
\varphi \in & \left] \frac{360 (n-1)}{N} , \frac{360n}{N} \right]
\end{align}

A pixel at $r'$, $\varphi'$ Polar coordinate is a member of $\vec{V}_{nm}$ segment if and only if the coordinates of the pixel fall in the boundaries of the segment. Accordingly, the membership of a pixel at $x$, $y$ Cartesian coordinate to $\vec{V}_{nm}$ segment depends on the following condition.
\begin{align} 
\tan^{-1}(\frac{x}{y}) \in & \left] \frac{360(n-1)}{N}, \frac{360n}{N} \right] \\
\vert \sqrt{x^2 + y^2} \vert \in & \left] \frac{m-1}{M}, \frac{m}{M} \right]
\end{align}

\subsection{Shape abstraction}
A shape is abstracted by aggregating pixels in the area of each segment. The binary representation of a shape enables the use of simple \textit{count} aggregate function that is the number of pixels represented by the segment with the value of $1$. Let $\gamma_{nm}$ denote the aggregated value of segment $\vec{V}_{nm}$ which is defined as $S_{nm} = \vert \{B_{xy} \vert B_{xy} = 1 \}  \vert$ for all $x$ and $y$ of pixels belonging to $\vec{V}_{nm}$ segment.

The proposed method operates upon $\gamma_{nm}$ only, and is independent from $x$ and $y$ components. Therefore, the segmentation matrix, $I$, is modified as it follows; and is called \textit{abstraction matrix} ($\Gamma$).
\[
\Gamma =
\bordermatrix{           & \text{Segment} \; 1  & \dots  & \text{Segment} \; M \cr
	  \text{Sector} \; 1 & \gamma_{11}          & \dots  & \gamma_{1M}         \cr
	  \dots              & \vdots               & \ddots & \vdots              \cr
	  \text{Sector} \; N & \gamma_{N1}          & \dots  & \gamma_{NM}}        \qquad
\]

The matrix is independent from the coordinates of each vector in 2D Euclidean space. However, the coordinates are implicitly approximated in the order of each of the vectors. For instance, $\gamma_{2m}$ refers to $m$-th segment on $2 \times (360/N)$-th sector. Finally the $\Gamma$ matrix is normalized using \textit{coefficient of variation} method.

\subsection{Translation} \label{section: Translation}
In most previous works such as Kabsch algorithm~\cite{kabsch1976solution}, translating input shapes to a position such that their centroid coincide with the center of coordinate system, or any specific coordinate, is a mandatory preprocessing. In general, translation superimposition is inevitable for both partial and full Procrustes superimpositions. 

The abstraction vectors of $\Gamma$ matrix are independent from $x$ and $y$ parameters, hence translation is not essential for the proposed method. However, an alternative application for translation is defined, which enables partial match determination between shapes. For this application, the translation process between the two input shapes could be interpreted as moving the segmentation center of one shape to coordinates pointed out by the segmentation vectors of the other shape (a process similar to translation superimposition in Procrustes analysis). 
Let $T_x$ and $T_y$ denote translation on $x$ and $y$ direction respectively; and $T=T_x \times T_y$ (Cartesian product of translations on $x$ and $y$ coordinates) be the set of all possible translations. The partial match between two shapes is determined by state-space search performed on the $T$ set, which is by applying all the transformations of $T$ on the second shape, and assessing the similarity between the first and second shape (see section~\ref{section: Similarity Measurement}).

\subsection{Rotation} \label{section: Rotation}
\begin{figure}[!t]
	\centering
	\includegraphics[width=\columnwidth]{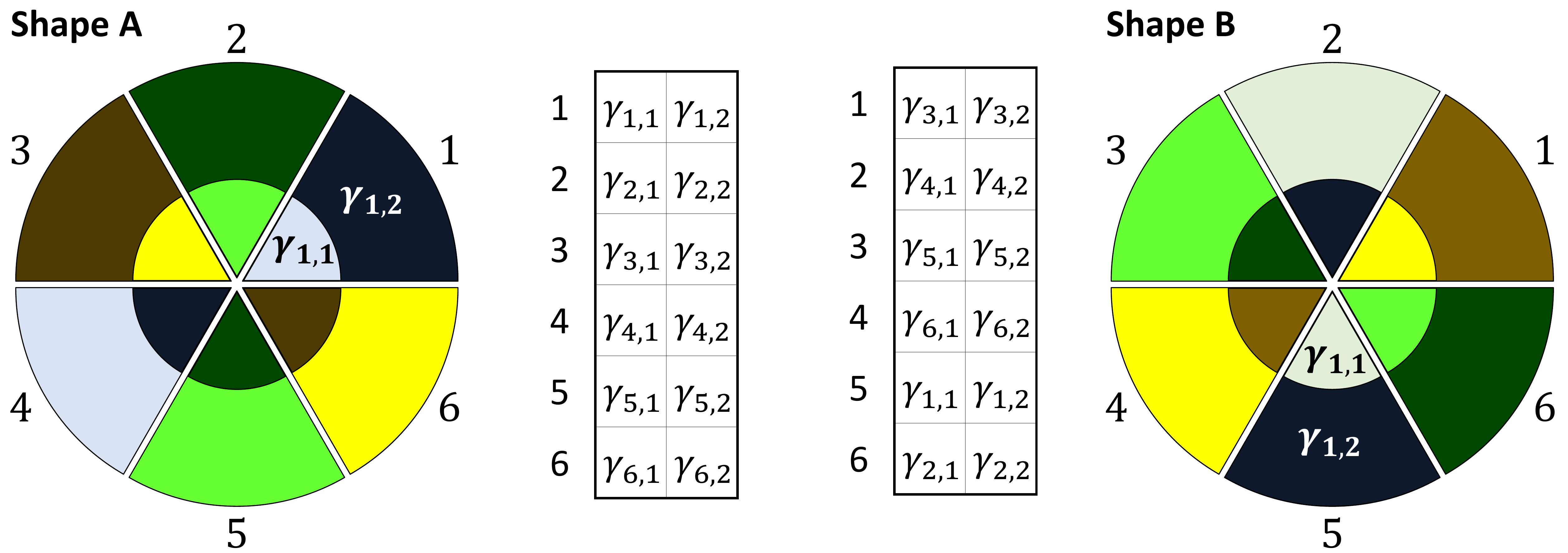}
	\caption
	{
		Abstractions of two inputs, \textit{Shape A} and \textit{Shape B}, are given with partitioning parameters $N=6$ and $M=2$. There is a $120^\circ$ rotation difference between the two shapes. Given $N$ and $M$, the set of rotation angles based on $(360/60)j$ is $R = \{0^\circ, 60^\circ, 120^\circ, 180^\circ, 240^\circ, 300^\circ, 360^\circ \}$. These rotations are implemented using circular shift on $\Gamma$. Accordingly, if the transformation between inputs is $(360/60)j$ degrees of rotation, it is superimposable by $j$ circular shifts of $\Gamma$. Therefore, four times circular shift on $\Gamma_B$ results highest similarity value (e.g., $J=1$), which yields $120^\circ$ rotation as the best linear mapping transformation between \textit{Shape A} and \textit{Shape B}.
	}
	\label{Figure: ShapeRotation}
\end{figure}

Rotation is a rigid body motion of a space that maintains at least one point at its original location; here we fix the segmentation center and move the segmentation vectors. In other words, given that a frame is partitioned into $360/N$ equal sectors, any $(360/N)j$ degrees of rotation for $j \in \{0, 1, \dots N-1\}$ is implemented as $j$ units of circular shift on $\Gamma$ (see Fig.\ref{Figure: ShapeRotation}).

Following the aforementioned objective of using basic operations, rotation is implemented using circular shift operation on $\Gamma$. Accordingly, given $N$ sectors (given that rotation is a rigid body transformation, this operation is independent from $M$), a set of rotation angles that are implemented using circular shift on $\Gamma$, is defined as it follows.
\begin{equation}
R = \left\{ \frac{360}{N}i \,\middle|\, i= 0, 1, \dots N-1 \right\}
\end{equation}

The set $R$ defines a discreet set of rotation angles which are hierarchically extended to a continuous domain using an iterative procedure discussed in Section~\ref{section: Iteration}.

\subsection{Similarity measurement} \label{section: Similarity Measurement}
To determine the best linear mapping transformation, the presented method runs a state-space search on transformations, by transforming the second shape, and assessing it's similarity with the first shape. In general, let $\Delta = T \times R$ be the Cartesian product of $T$ translation and $R$ rotation. Note that, if the alternative application for translation introduced in Section~\ref{section: Translation} is of no interest, thence $\Delta = R$. The linear mapping transformations between \textit{Shape A} and \textit{Shape B}, are ranked based on the similarity coefficients $J(\Gamma_A, \delta\Gamma_B)$ for $\delta \in \Delta$.

The similarity between any two elements $\gamma_{nm}^A \in \Gamma_A$ and $\gamma_{nm}^B \in \Gamma_B$ is measured using \textit{Jaccard similarity coefficient}, denoted $j(\gamma_{nm}^A, \gamma_{nm}^B)$, and is calculated as it follows.
\begin{equation}
j(\gamma_{nm}^A, \gamma_{nm}^B) = \frac{\vert \gamma_{nm}^A - \gamma_{nm}^B \vert}{\gamma_{nm}^A+\gamma_{nm}^B}
\end{equation}
The similarity between two abstracted shapes, denoted $J(\Gamma_A, \Gamma_B)$, is calculated as the sum of all pairwise Jaccard similarity indexes as it follows.
\begin{equation}
J(\Gamma_A, \Gamma_B) = \sum_{nm} j(\gamma_{nm}^A, \gamma_{nm}^B)
\end{equation}

Tanimoto similarity coefficient~\cite{rogers1960computer} is an alternative to Jaccard index; however since both methods yield similar results, Jaccard index is chosen. Additionally, other alternatives to Jaccard index are: S{\o}rensen similarity index~\cite{sorensen1948method}, Bray–Curtis dissimilarity~\cite{bray1957ordination} (also known as Czekanowski similarity index), Pearson product moment correlation, and earth mover's distance~\cite{rubner2000earth}.

The similarity assessment between any two $\gamma_{nm}^A$ and $\gamma_{nm}^B$ is also optionally extended by a neighborhood operation. Let $j(N_k(nm, i))$ denote the Jaccard index of neighbor $N_k$ of element $n$, $m$ at $i$-th distance. The extended similarity coefficient $j'$ is calculated as it follows for $d$ neighbors:
\begin{equation}
\begin{split}
j'(\gamma_{nm}^A, \gamma_{nm}^B) =& J(\gamma_{nm}^A, \gamma_{nm}^B)\\
+& \sum_{i=1}^{d} \left[ \log_{d+2} (d+2-i) \sum_{K_i} j(N_k(nm, i)) \right]
\end{split}
\end{equation}

\noindent
This is an adaptive logarithmic neighborhood operation which assigns heavier weight to closer elements than remotes. 

\subsection{Iteration} \label{section: Iteration}
How long does it take us to understand the gist of a shape? Henderson et al.~\cite{henderson1998eye} obtained a typical scene fixation of 304ms with 100\% luminance, and Rayner~\cite{rayner1998eye} estimated $233$ms fixation time for an adult reading normal text (Kowler et al.~\cite{kowler1987reading} measured fixation patterns for reading reversed letters). The conceptual and perceptual information understood from a glance at an image frame is a function of the glance duration. Fei-Fei et al.~\cite{fei2007we} studied the perception depth over time. He resulted that we perceive sensory information (e.g., dark and light) in roughly $50$ms, at $107$ms we determine more semantic aspects (e.g., people, room, urban, and water) with considerable accuracy, it takes $150$ms to determine an object (e.g., dog), and at $500$ms we achieve maximum perception (e.g., identify dog as German shepherd).  Greene et al.~\cite{greene2009briefest} conducted similar study and established a perceptual benchmark to types of information we perceive during early perceptual processing; and they inferred that it takes $63$ms to determine naturalness of an image frame and $78$ms to understand whether its forest or not. Such global-to-local view cognitive abilities inspired the present study to see transformation determination as a multi-step procedure an opposed to the some single-step methods.

The longer we are exposed to a shape, the more we understand from it; in other words, the amount of information we perceive from a shape is the function of the number of image processing iterations performed on the shape. Accordingly, present study defines a converging heuristic iterative method that determines the gist of shapes at initial approximation (highest abstraction that corresponds to sensory information such as light/dark classification~\cite{fei2007we}), and by translation superimposition followed by similarity assessment procedure, the most abstracted transformation is determined. Then, segmentation parameters are iteratively incremented. At each iteration, a new abstraction with more details than its preceding abstraction is made. Also, at each iteration, approximated transformations of the preceding iteration are tuned using more detailed abstraction. This process is analogous to: from light, through animal and dog, to German shepherd~\cite{fei2007we,greene2009briefest}.

In general, the proposed method determines an initial approximation of best mapping transformations, and tunes those through successive iterations. The permutations of transformations at each iteration form a state-space that is traversed in best-first search fashion. This approach follows the traits of \textit{Greedy algorithm}~\cite{cormen2009introduction} that determines local optimal choice. To best of our knowledge, cognitive community descriptions of processing segmentations, well overlaps local optimal search method. However, one may consider updating the procedure to follow traits of global optimal search methods to best adapt the application requirements.

Let $l \in \mathbb{N}$ denote an iteration coefficient which is initialized with a user-defined parameter $\omega$. Let $\Gamma_A^l$ be abstraction of shape $A$ at iteration $l$ with $N_l = 2^l$ sectors and $M_l=2^l$ segments. For the purpose of readability of the method, the number of sectors and segments are chosen to be identical; however, the extension of the method to support divers parameters is straight-forward. According to this generalization, the amount of details represented by each abstraction grows exponentially through the iterations. Also, the growth rate can be update to best adapt the application requirements by changing either the growth function or considering $l\in \mathbb{R}$.

Let $\Delta_l$ be the set of all transformations to be applied on $\Gamma_A^l$ at iteration $l$ for $\Delta_\omega = T \times R$. Let $\Upsilon_l = \{\delta_1 \dots \delta_i \dots \delta_\epsilon \}$ be the set of top-$\epsilon$ transformations (i.e., highest similarity) at iteration $l$ for $\epsilon$ being a user-defined parameter. The iteration $l$ tunes best transformations of iteration $l-1$. Accordingly, $\Delta_l$ consists of all $\Upsilon_l$ tunes which is formally defined as it follows for the user-defined parameter $\lambda$ that specifies the tuning range. 
\begin{equation}
\forall j \in \{0, 1, \dots \lambda\} \colon \Delta_l = \{(2^{l-\omega} \delta_{(l-1)i} ) \pm j \}
\end{equation}

For instance, suppose $\omega=3$ then $N=8$ and $M=8$ and assuming only rotation superimposition, we obtain $\Delta_3 = \{0,1,2, \dots 7\}$ which are the number of circular shifts on $\Gamma_A$ that corresponds to $\{0^\circ, 45^\circ, 90^\circ, \dots 315^\circ\}$. Suppose $\epsilon=1$, $\Upsilon_3={2}$, and $\lambda=2$, accordingly $\Delta_4$ is calculated as it follows.
$\Delta_4=\{(2^{4-3} \times 2) \pm \{0, 1\}\}$

\noindent
which corresponds to $\{67.5^\circ, 90^\circ, 112.5^\circ\}$. The pseudo code of the iteration procedure is given in Algorithm~\ref{Algo: IterationAlgorithm}.

\begin{algorithm}
	\caption{Iteration Algorithm}\label{Algo: IterationAlgorithm}
	\begin{algorithmic}[1]
		\Procedure{Iterate}{}
		\State $l \gets \omega$
		\State $\Delta_l \gets T \times R$
		\State $N_l \gets 2^l$
		\State $M_l \gets 2^l$
		\State \textbf{Build} $\; \Gamma_A^l \;$ and $\; \Gamma_B^l \;$
		\State $\Upsilon \gets \text{apply} \; \Delta_l \; \text{on} \; \Gamma_A^l \; \text{and get top-}\epsilon \; \text{transformations}$
		\If {$l < \max l$}
		\State $l \gets l+1$
		\State $\Delta_l \gets \text{all tunes of} \; \Upsilon_{l-1}$
		\State \textit{Goto} 03
		\Else
		\State report $\;\Upsilon_l\;$ as best mapping transformations
		\EndIf
		\EndProcedure
	\end{algorithmic}
\end{algorithm}

\subsection{Validation and verification} \label{section: Validation}
If the difference between two input image frames is $\theta^\circ$ and $\theta \in \Delta_\omega$, according to Section~\ref{section: Rotation}, then $\theta$ is determined using circular shifts on $\Gamma$. However, if $\theta \notin \Delta_\omega$, then $\theta$ is determined using the iterative process. To prove that, consider the following hypothesis:
\begin{equation}
\exists n \in \{1, 2, \dots N \} \colon \frac{360}{N}(n-1) < \theta \leq \frac{360}{N}n 
\end{equation}

\noindent
By definition of segmentation, these are the boundaries of $n$-th region which is divided into $M$ equal segments. According to the hypothesis, $\theta$ belongs to one and only one region, therefore as much as the range is narrowed-down, we get closer and closer to $\theta$ (the motivation of iteration procedure).
At each iteration, the results of former iteration are tuned until a result with a user-defined accuracy ($\rho$) is determined. The algorithm performs maximum $c$ iterations which is calculated as it follows.
\begin{itemize}
	\item The segmentation area should be as narrow as $\rho^\circ$, therefore: 
	\begin{equation}
	\rho = \frac{360}{N}n- \frac{360}{N}(n-1) \rightarrow \rho = \frac{360}{N}
	\end{equation}
	
	\item Segmentation grows exponentially through iterations, hence:
	\begin{equation}
	N_l = 2^l \rightarrow \frac{360}{\rho} = 2^l \rightarrow l = \log_2 \frac{360}{\rho} \rightarrow l = \lceil \log_2 \frac{360}{\rho} \rceil
	\end{equation}
\end{itemize}

\section{Results}
The accuracy of proposed method is assessed using $100$+ pairs of image frames with diverse resolutions spanning from $50 \times 50$ to $1000 \times 1000$ pixels, and including different categories (e.g., animals, cars, airplane, people, and abstract images). Additionally, the impact of noisy image frames to the accuracy of the proposed method, is assessed using image frames with up to $70$\% of random noise

The evaluations are designed as it follows
\begin{inlinelist}
	\item \textit{Shape A} is a BMP image, or an abstract shape composed of lines, circles, and random noise drawn using features integrated in the implemented tool
	\item \textit{Shape B} is obtained by $\theta$ degrees rotation of \textit{Shape A} plus a random percentage of noise
	\item \textit{Shape A} and \textit{Shape B} are used as inputs for the proposed method
	\item the central tendency of determined rotations is measured as weighted arithmetic mean among top-3 (WM3) rotations, and it is compared with $\theta$. 
\end{inlinelist}
The assessments are performed with default parameters which are: $\omega = 3$, $\lambda=10$, $\epsilon=10$, and the similarity between two segments is calculated excluding neighbor segments. The results of the experiments are discussed as it follows.

\subsection{Top transformations converge rapidly}
The fundamental argument of iterations is to progressively increase the level of details on the image frame abstraction, and accordingly, iteratively improve the accuracy of the calculated approximated transformations, until a user-defined precision criterion is met. Weighted sample variance among top-3 (WV3) approximated transformations provides a measure of dispersion on top approximations. The WV3 reflects the variability in the top-3 approximated transformations, such that: a small WV3 suggests a very reliable WM3, while a large WV3 reflects an uncertainty about the ``best'' linear mapping transformation. According to the experiments, WV3 gets closer to $1$ in a few iterations which yields (a) rapid convergence among top approximated transformations (this confirms the validation of iteration procedure discussed in Section~\ref{section: Validation}), (b) $\textit{WV3} \approx 1$ in few iterations ($>6$) confirms the accuracy of rapidly converged approximated transformations.

\begin{figure*}[!ht]
	\centering
	\includegraphics[width=0.9\textwidth]{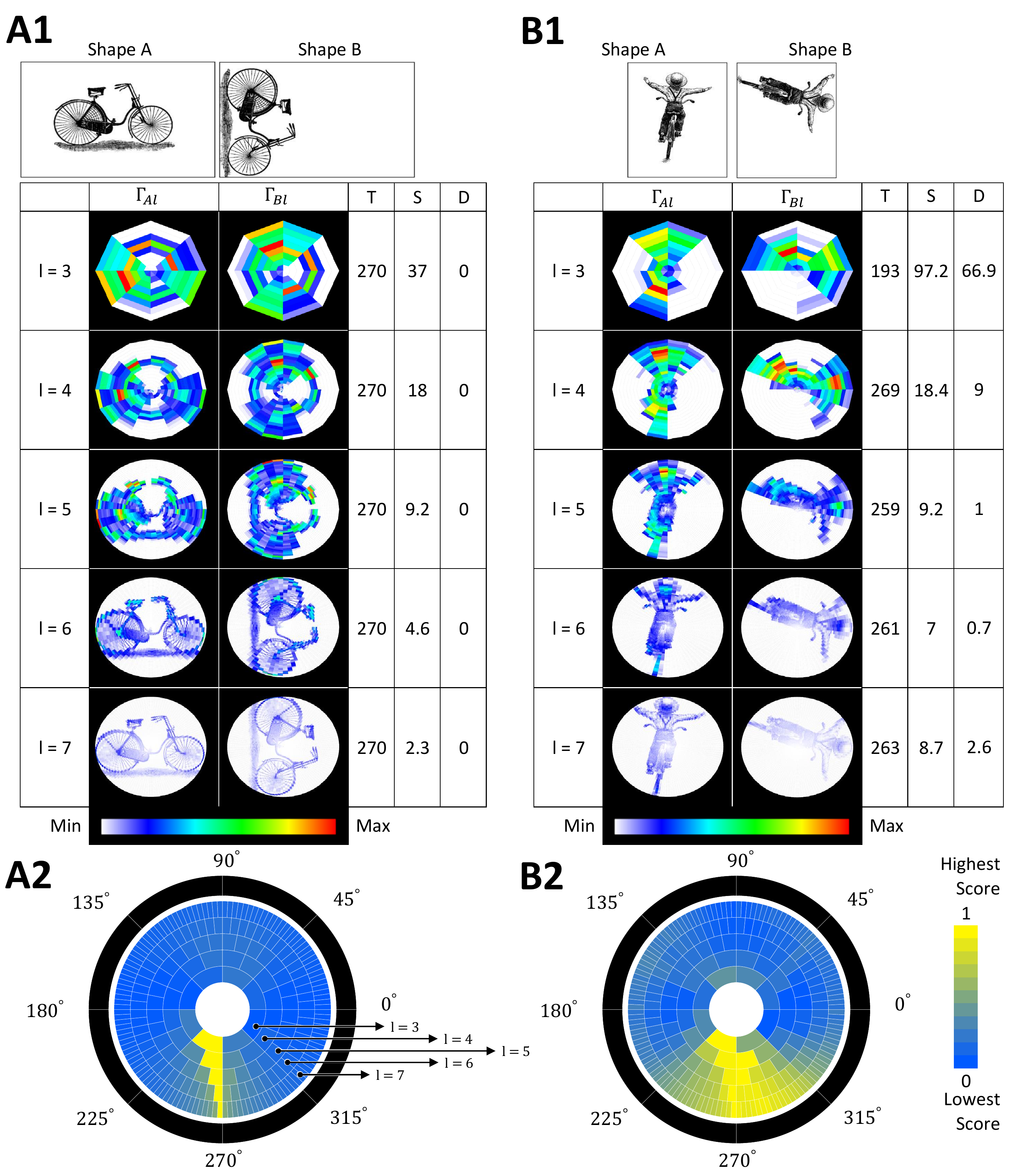}
	\caption
	{
		\textit{Shape A} is loaded from a BMP image, and \textit{Shape B} is obtained by $270^\circ$ rotation of \textit{Shape A}. The $\Gamma$ matrices of both shapes at different iterations are presented by circular heatmaps. T: determined transformation, S: standard deviation among top-3 determined transformations, D: difference between actual and determined transformations. The normalized similarity index $J(\Gamma_A, \delta \Gamma_B), \forall\delta \in \Delta$ is plotted using a circular heapmap for all the iterations, see panels A2 and B2.  
	}
	\label{Figure: 270}
\end{figure*}

\subsection{Tuning out the cognitive noise}
Selective and visual attention filter irrelevant stimuli to the subject's task by mechanisms such as habituation and cognitive inhibition. There have been promising efforts to model the ability (e.g.,~\cite{tsotsos1995modeling}) since the \textit{spotlight}~\cite{eriksen1972temporal} and \textit{zoom lens}~\cite{eriksen1986visual} models. Additionally, perceived visual information are function of an observer's distance to an object. This aspect has variety of applications namely is Olivia et al.~\cite{oliva2006hybrid} that incorporates this aspect with hybrid images. A hybrid image is composed of two image frames with low and high spatial frequencies, such that either is perceived as noise as a function of observer's distance to the hybrid image frame. In other words, the image of high spatial frequency is dominant at closer distance, while the image with low frequency is perceived at far distance. Whether the noise is a masked image or it is an irrelevant stimuli, it does not impact the perceived information from an image frame. Therefore, the performance of proposed method in approximating linear mapping transformation using noisy image frames, is assessed by experiments where a percentage of \textit{Shape B} is covered with random noise. 

\begin{figure}[!t]
	\centering
	\includegraphics[width=\columnwidth]{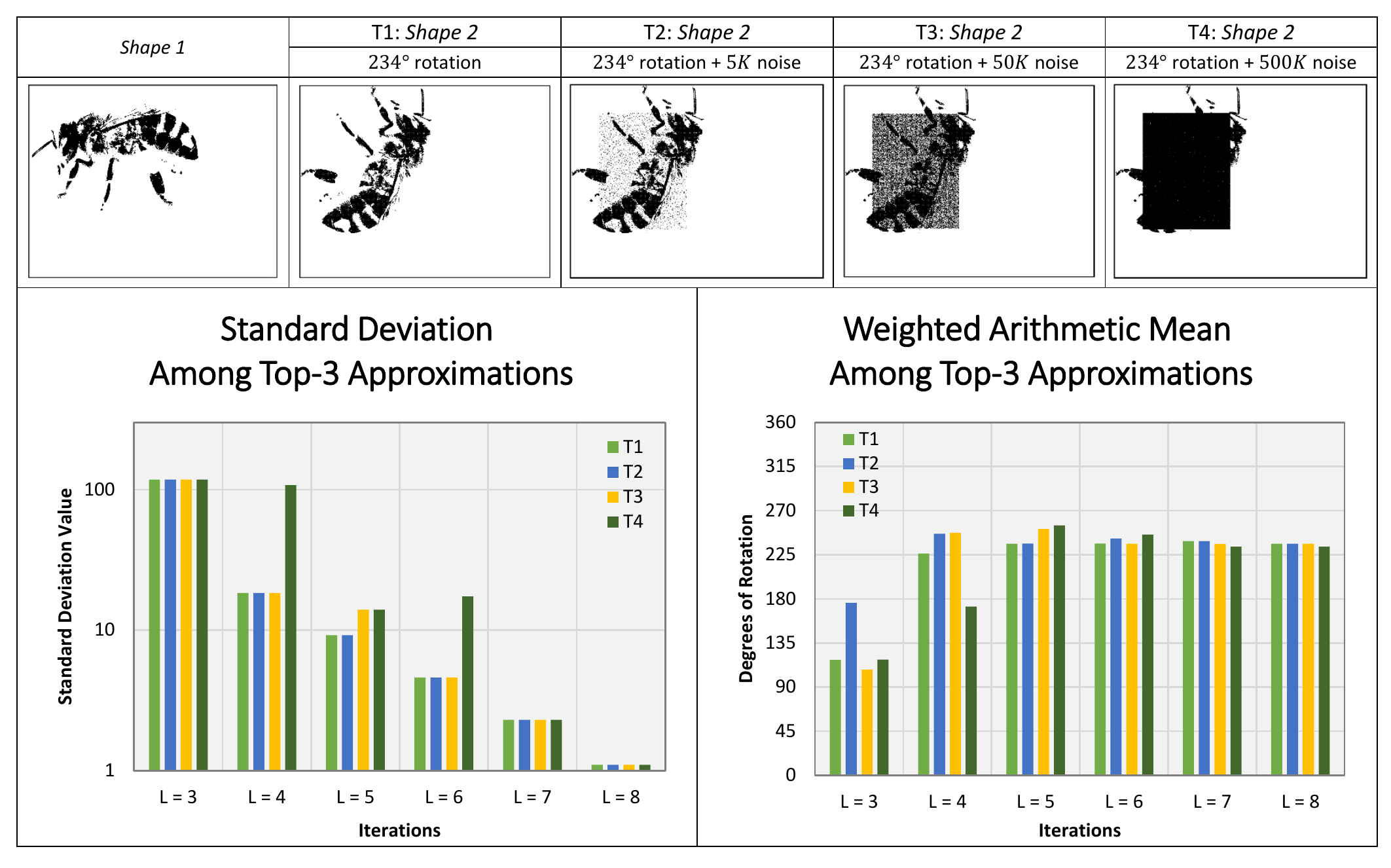}
	\caption
	{
		Evaluation of random noise impact on transformation determination.s
	}
	\label{Figure: NoiseImpact}
\end{figure}

To this extend, an experiment of four tests, $T1$, $T2$, $T3$, and $T4$ is conducted (see Fig.~\ref{Figure: NoiseImpact}). The tests have \textit{Shape A} in common which is a BMP image of a bee. The \textit{Shape B} is created by $234^\circ$ rotation of \textit{Shape A}, and differs among test in the amount of incorporated random noise. The subject in the \textit{Shape A} (i.e., the bee) is represented by $\approx230$K pixels (of $584$K pixels of the image frame). A portion of $120$K pixels (out of the $\approx230K$ pixels) is subject to random noise. This portion is intentionally chosen to cover the body of the bee which presents the majority of perceptible features of the subject. Given that the pixels are binary and the figure is represented by pixels of value 1 (see Section~\ref{section: Shapre Representation}), the random noise is created by setting the value of a random pixel to $1$ in the subject-to-noise portion of \textit{Shape B}. The random noise is added through an iteration of $0$, $5$K, $50$K, and $500$K random pixel selections (a pixel can be selected multiple times) respectively for $T1$, $T2$, $T3$, and $T4$ (see Fig.~\ref{Figure: NoiseImpact}); such that, the majority of perceptible features on \textit{Shape B} are covered with random noise at $T4$. 

The initial segmentation parameter ($\omega = 3$) provides a limited number of variant initial approximations (see Section~\ref{section: Iteration}). Therefore, the WV3 at first iteration (i.e., $l=3$) of the $T1$, $T2$, $T3$, and $T4$ show relatively high dispersion, which indicate the inconsistency of WM3 (see Fig.~\ref{Figure: NoiseImpact}). The initial approximations are tuned at second iteration (i.e., $l=4$) which improve WV3 tenfold (from $118$ to $18$) for the $T1$, $T2$, and $T3$. Despite of a minor discrepancy, WM3 of the tests $T1$, $T2$, and $T3$ are relatively close to actual transformation (i.e., $234^\circ$). However, the considerable noise of $T4$ prevents its WV3 convergence at the same rate as of $T1$, $T2$, and $T3$ (see Fig.~\ref{Figure: NoiseImpact}). The third iteration (i.e., $l=5$) improves approximations, and it brings WV3 of all the test to a same scale, and accordingly provides reliable WM3. Further iterations squeeze the approximations and reach to $\text{WV3}=1.1$ for all tests at sixth iteration (i.e., $l=8$) which indicates a considerable consistency of WM3. Therefore, the method determines WV3 and WM3 for all test at the same scale, given the considerable amount of noise (specially at $T4$). This confirms that even a low amount of perceptible features of the figures is adequate to tune the initial approximations to reliable approximations. For details of the noise impact on other approximations, refer to Supp. Fig.2.17-20.

\subsection{Image resolution defines maximum number of iterations}
When abstracting an image frame, up until a certain iteration, a segment consists of multiple pixels. However beyond that iteration, a segment might be smaller than a pixel (i.e. one pixel belongs to multiple segments). To determine a segment to which a pixel belongs to, the method rounds the position of the pixel. Therefore, beyond a certain iteration, the rounding procedure could potentially increase the distance between the abstractions of two image frames. In such condition, the WV3 converges up-until a certain iteration, and it is saturated beyond that iteration, and accordingly is the WM3 (see Supp. Fig.2.22-23). Therefore, maximum number of iterations, and accordingly the number of \textit{segments} and \textit{sectors} are the function of shape resolution.

\subsection{Pin-pointed transformation vs. condensed approximations}
The linear mapping transformation between two shapes is determined either as a single transformation with considerable discrepancy with the rest of the approximations (e.g., panel A on Fig.~\ref{Figure: 270}), or a condensed distribution of approximated transformations around actual transformation (e.g., panel B on Fig.~\ref{Figure: 270}). This behavior originates from the discreet representation of image frames (raster graphics); such that, when drawing a \textit{Shape B} from \textit{Shape A}, a pixel of \textit{Shape A} is mapped to a rounded position on \textit{Shape B}. Therefore, pixels of \textit{Shape A} could overlap as mapped on \textit{Shape B}. For instance, the two pixels at $\langle x_1=4, y_1=4 \rangle$ , $\langle x_2=4, y_2=5 \rangle$ belonging to the segment/sector $V_{nm}$ of \textit{Shape A}, with $70^\circ$ rotation, respectively map to positions $\langle x_1^\prime = -0.562, y_1^\prime = 5.628 \rangle$ and $\langle x_2^\prime = -1.33, y_2^\prime = 6.262 \rangle$. As the coordinates are rounded, the two pixels map to position $\langle -1, 6 \rangle$ belonging to the segment/sector $V_{n'm'}$ of \textit{Shape B}. Therefore, two pixels of \textit{Shape A} map to one pixel on \textit{Shape B} (surjective linear transformation). Accordingly, as abstracting the shapes using aggregation function \textit{count} (see Section~\ref{section: Shape Segmentation}), the abstraction parameters are calculated as it follows: $\gamma_{nm} = 2$ and $\gamma_{n'm'} = 1$ (e.g., see comparison of $\gamma$ value distribution plots on Supp. Fig.2.1-22). Hence, comparing $\gamma_{nm}$ and $\gamma_{n'm'}$ results to $j(\gamma_{nm}, \gamma_{n'm'}) = 0.33$ as opposed to expected $j(\gamma_{nm}, \gamma_{n'm'}) = 1$. Such scenarios prevents ``pin-pointing'' the actual transformation (in this case $70^\circ$) and rather provides a condensed distribution of transformations around actual transformation (e.g., see panel B on Fig.~\ref{Figure: 270}).

\subsection{A small similarity is sufficient to determine a reliable linear mapping approximation}
Ideal scenario for comparing two shapes is when there exist a one-to-one correspondence (injective/surjective) between pixels of two the shapes. However, for variety of reasons discussed as it follows, the rotation function on raster graphics is surjective. For instance, rotation function may map multiple pixels of \textit{Shape A} to one pixel of \textit{Shape B}, causing a percentage of deformation on \textit{Shape B} (e.g., see supp. Fig.2.21), and preventing ``pin-pointing'' actual transformation (as above-discussed). Additionally, shapes are possibly subject to noise, which would prevent one-to-one correspondence between the two shapes (non-surjective). Moreover, \textit{Shape A} may consist of congruent figures (e.g., two side-by-side circles of the same radius), and if \textit{Shape B} is determined by $\theta^\circ$ rotation of \textit{Shape A}, then in addition to $\theta^\circ$, multiple rotation angles may also map the congruent shapes on each other. In such cases, actual transformation is determined using incongruent elements (e.g., saddle area, or pedal of the bicycle on Fig.~\ref{Figure: 270}). Such prominent details not only improve approximations for congruent shapes, but are also advantageous when the majority of the shape is covered by noise (e.g., Fig.\ref{Figure: NoiseImpact}) or is deformed (e.g., Supp. Fig.2.21). 

The method discussed in present study, minimizes the impact of such discrepancies on linear mapping transformation determination, by calculating the similarity of two corresponding segments independently from the rest of the segments (an adaptive neighborhood operation of custom range is optionally enabled). Therefore, a higher similarity between few segments is adequate to determine mapping transformation with considerable accuracy. The experiments on deformed, congruent, and noisy image frames illustrate the accuracy of the proposed method on such scenarios.

\section{Conclusion and Discussion}
Understanding human brain mechanism and adapting it to different disciplines has always been of interest in various disciplines, such as in deep learning~\cite{hinton2007learning,hinton2006fast}. Ballard defined deictic computation~\cite{ballard1997deictic} that benefits computational methods to study the connection between body and real world movements, to cognitive tasks. Roger B. Nelsen’s ``Proofs without Words''~\cite{mackenzie1993proofs} and Martin Gardner’s ``aha! Solutions''~\cite{gardner1978aha} encouraged the present research to focus on a cognitive-oriented approach for linear mapping transformation between two shapes. The proposed method iterates over different abstractions of image frames, from the most abstracted to the most detailed, while tuning the top transformations of previous iteration to obtain best mapping linear transformations.

The proposed method is implemented and tested over variety of inputs. The method is assessed for its accuracy on determining linear mapping transformations. The experiments showed that the output of the method is reliable even under challenging conditions such as deformed and noise image frames. Additionally, the size of abstraction matrix ($\Gamma$) is independent from the size of the input images frames, and accordingly the computational cost of the proposed method is independent from the resolution of input image frames.

\ifCLASSOPTIONcaptionsoff
  \newpage
\fi

\bibliographystyle{IEEEtran}
\bibliography{IEEEabrv,Biblio}

\end{document}